# PFL-LSTR: A PRIVACY-PRESERVING FRAMEWORK FOR DRIVER INTENTION INFERENCE BASED ON IN-VEHICLE AND OUT-VEHICLE INFORMATION

**Runjia Du**
Graduate Research Assistant, Lyles School of Civil Engineering, Purdue University, West Lafayette, IN, 47907
Email: du187@purdue.edu

**Pei Li**
Scientist, Department of Civil and Environmental Engineering, University of Wisconsin-Madison, Madison, WI, 53706
Email: pei.li@wisc.edu

**Sikai Chen***
Assistant Professor, Department of Civil and Environmental Engineering, University of Wisconsin-Madison, Madison, WI, 53706
Email: sikai.chen@wisc.edu
(Corresponding author)

**Samuel Labi**
Professor, Lyles School of Civil Engineering, Purdue University, West Lafayette, IN, 47907.
Email: labi@purdue.edu

**ABSTRACT**
Intelligent vehicle anticipation of the movement intentions of other drivers can reduce collisions. Typically, when a human driver of another vehicle (referred to as the target vehicle) engages in specific behaviors such as checking the rearview mirror prior to lane change, a valuable clue is therein provided on the intentions of the target vehicle's driver. Furthermore, the target driver's intentions can be influenced and shaped by their driving environment. For example, if the target vehicle is too close to a leading vehicle, it may renege the lane change decision. On the other hand, a following vehicle in the target lane is too close to the target vehicle could lead to its reversal of the decision to change lanes. Knowledge of such intentions of all vehicles in a traffic stream can help enhance traffic safety. Unfortunately, such information is often captured in the form of images/videos. Utilization of personally identifiable data to train a general model could violate user privacy. Federated Learning (FL) is a promising tool to resolve this conundrum. FL efficiently trains models without exposing the underlying data. This paper introduces a Personalized Federated Learning (PFL) model embedded a long short-term transformer (LSTR) framework. The framework predicts drivers' intentions by leveraging in-vehicle videos (of driver movement, gestures, and expressions) and out-of-vehicle videos (of the vehicle's surroundings – frontal/rear areas). The proposed PFL-LSTR framework is trained and tested through real-world driving data collected from human drivers at Interstate 65 in Indiana. The results suggest that the PFL-LSTR exhibits high adaptability and high precision, and that out-of-vehicle information (particularly, the driver's rear-mirror viewing actions) is important because it helps reduce false positives and thereby enhances the precision of driver intention inference.

**Keywords:** Human driver intention, Deep Learning, Transformer model, data privacy, Personalized Federated Learning.

## INTRODUCTION

Of the 40,000 U.S. highway deaths in 2009, 18% were attributed to lane control issues (e.g., merging, changing lanes, negotiating curves, and overtaking) (*1*). Crashes are often associated with these maneuvers because they inherently require awareness of the neighboring vehicle of the ego vehicle's intentions, otherwise, those other vehicles are left unprepared to respond to the ego vehicle's maneuvers. Studies suggest that, on average, drivers require approximately 2.3 seconds to react, making it challenging to prevent the collision (*2*). The occurrence of these traffic crashes can largely be attributed to human error, accounting for approximately 90% of these incidents (*3*). Vehicle automation has been widely regarded as a promising solution for minimizing traffic accidents, owing to its rapid development. By eliminating human factors from vehicle control through automation, the potential to decrease traffic crashes is greatly enhanced (*4*). Despite the rapid progress in intelligent vehicle technology, transportation vehicles of the current era are still several years or decades from achieving full connectivity and automation(*5–7*). For this reason, the specter of human error will continue to persist in mixed traffic environments (that consist of automated vehicles and human-driven vehicles (*8–10*)). The statements above underscore the importance of future vehicles acquiring a more comprehensive understanding of the intentions of human drivers.

In this paper, a "target vehicle" refers to the vehicle intending to change lanes, characterized by specific driver behavior. To identify this intention, three cues are commonly used: turn signal activation, vehicle dynamics analysis (lateral speed and steering angle), and observation of driver behavior (head rotation and eye movement) (*11*). Nonetheless, only 66% of lane changes involve the utilization of turn signals, often occurring within a second of, or even after, the commencement of the lane change (*12*). The identification of indicators before initiating driving maneuvers is vital. Two such indicators that provide valuable information prior to the maneuvers are: (a) the driving behaviors, and (b) the surrounding traffic environment. It has been estimated that 65% to 92% of drivers check rear mirrors seconds before changing lanes (*13*). Moreover, as mentioned by Ou et al., drivers' observation actions in observation periods seek to decide whether to undertake an action (*14*). For example, if the vehicle in front is closely following the target vehicle, it indicates that the current lane's travel speed is being influenced, potentially leading to a lane-change behavior Figure 1. Similarly, if there exist vehicles (in the adjacent lane) in close proximity to the target vehicle, the driver may opt to abort the lane-change maneuver due to safety concerns. Therefore, it's important to fuse information from different sources. However, it is important to note that vehicle related data often contain sensitive information, including facial features, vehicle details, geolocation information, and surrounding vehicle info (*15*). Therefore, it is of utmost importance to carefully address the preservation of user privacy for all vehicles on the road.

Federated Learning (FL) presents a solution that enables devices to train and share parameters with a server securely without the need to transmit data to the cloud (*16*). This approach effectively safeguards users' privacy and has found application in diverse fields, including healthcare, smartphone services, and more (*17, 18*). Despite its privacy-preserving benefits, one limitation of Federated Learning (FL) is that it creates a common model among all clients or users, which could pose challenges in tailoring the model to individual user data (because tying a set of actions to a specific individual is more complex within the FL framework). Fortunately, there exist various approaches to incorporate such personalization into the traditional Federated Learning (FL) framework, known as Personalized Federated Learning (PFL). PFL methodologies enable customization and adaptability to individual user preferences,

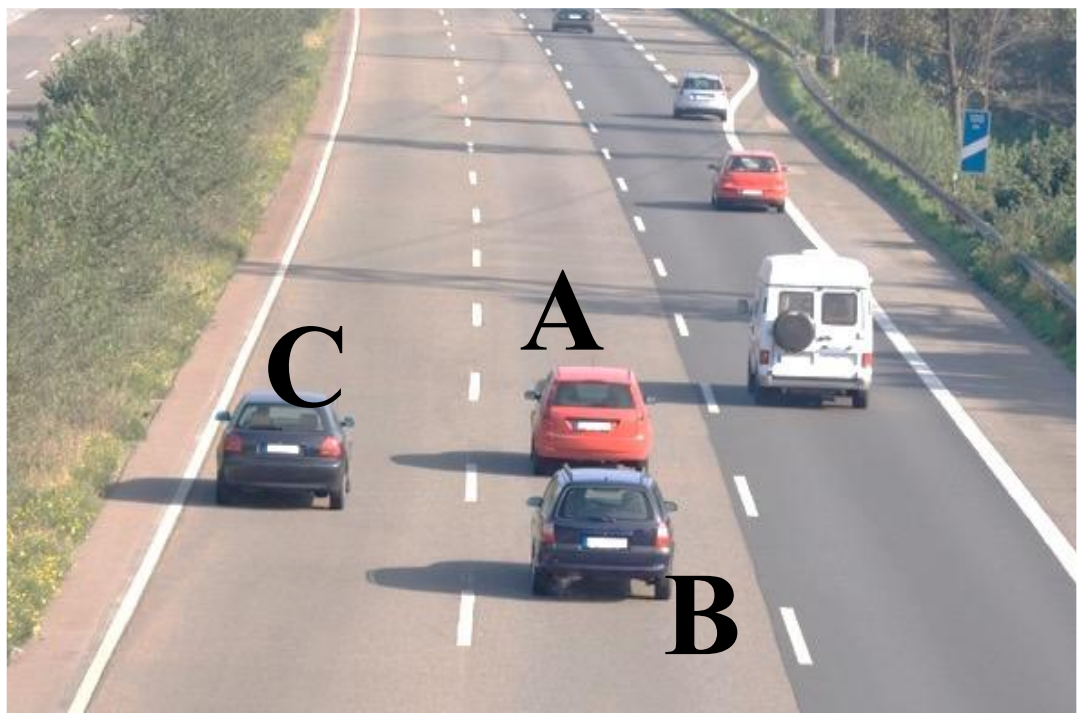

**Figure 1. Vehicle A's close proximity to B might trigger B to change lane to the left lane (target lane); however, close proximity of vehicle C may cause Vehicle B to abort the lane change**

addressing the limitation of the vanilla FL framework. Classic ways to do this include: (a) local fine tuning, which is the dominant approach for personalization. Each client receives a global model and tune it using its own local data and several gradient descent steps. This approach is predominantly used in meta-learning methods such as MAML (*19*), domain adaption or transfer learning (*20*–*22*); (b) multi-task learning, which takes optimization on each client as a new task. Also, cluster groups of clients based on some features such as region and user info (*23*). However, this might violate user privacy.

In the realm of driving maneuver prediction or driver intention inference tasks, initial approaches utilized first-generation machine learning methods, such as Support Vector Machines (SVM) and Relevance Vector Machines (RVM), which demonstrated favorable outcomes. Subsequently, to capture the interplay of temporal dependencies within sequences and long-term dependencies, neural network models such as Recurrent Neural Networks (RNN) and Long Short-Term Memory (LSTM) were introduced(*7, 24*). In recent years, the Transformer model has garnered substantial popularity in this field, owing to its remarkable capacity to capture contextual relationships within sequential data. Moreover, the Transformer model has gained popularity over LSTM due to its ability to parallelize processing, capture long-range dependencies, handle long sequences effectively, and provide interpretability through attention mechanisms. Long Short-term TRansformer (LSTR) model that brought up by Xu et al., jointly models long- and short-term temporal dependencies (*25*). Two main advantages are: (a) it stores the history directly thus avoiding pitfalls of recurrent models, and (b) it separates long- and short-term memories, which allows modeling short-term context while extracting useful correlations from long-term history. Thus, using LSTR in driver intention inference tasks seems to be appropriate and adequate.

In this study, we focus on improving the performance of driving intention inference. Unlike existing approaches that overlook personalization, user privacy, and false positive cases this study addresses these issues by combining comprehensive in-vehicle driver and out-of-vehicle surrounding information. It is worth noting that lane-keep intentions are sometimes erroneously classified as lane-change intentions. Our approach incorporates a fine-tuning based Personalized Federated Learning (PFL) framework with the LSTR model. This allows the creation of a more accurate system for predicting drivers' steering actions with high precision. By combining PFL with the long short-term memory section in LSTR, we can learn a personalized model with both common and unique representations. Our approach is evaluated on a diverse driving dataset, collected under various environments with different drivers. PFL-LSTR framework outperforms other baseline models such as FedAvg and local models under most cases on different maneuvers (lane-keep, lane-change on both directions). The contribution of this paper is two-fold:

1. We offer a combination of driving cues for predicting driver intentions, with a specific focus on utilizing rear information. This is particularly beneficial in situations where the driver (of the target vehicle) exhibits movement but does not engage in subsequent maneuvers.
2. We introduce a novel PFL-based LSTR model for predicting driver intentions. This model offers notable advantages, including high learning efficiency and enhanced adaptability. By leveraging the benefits of PFL, our proposed approach demonstrates improved performance and flexibility in accurately predicting driver intentions.

The remaining sections of this paper are structured as follows: Section II provides a review of related work and outlines the contributions of this paper. Section III presents a detailed introduction to the PFL-LSTR model framework. Section IV covers the experiment settings and includes various testing scenarios. Section V presents the results, followed by a discussion in the conclusion (Section VI).

## RELATED WORK

Driver intention inference is a vital research area, driven by the need for transportation safety. Anticipating potential behaviors allows vehicle assistance systems to prepare for upcoming events, enhancing road safety. Accurate inference relies on integrating diverse information, including surrounding traffic, vehicle's states, the driver's behaviors, and other relevant factors. Machine learning methods, utilizing various data types such as vehicle status, driving behaviors, and surrounding traffic information, play a crucial role in driver intention inference studies.

Tran et al., proposed an approach to predicting driver's intentions through steering wheel data using Hidden Markov Model (HMM), the driver performs maneuvers including stop/non-stop, change lane left/right and turn left/right in a simulator in both highway and urban environments(*26*). By considering a mixed-traffic scenario where both autonomous vehicles and human driving vehicle exist, Liu et al., proposed a driving intention inference framework based on HMM using vehicle mobility features (*27*). Since driver intention such as lane-change is not only related to the dynamic parameters of the vehicle, but also is closely related to the complex and changing external environment, Li et al., proposed different decision factor used in the HMM model through the driver lane change decision-making analysis, which reflects the influence of the external environmental on driver lane changing operation (*28*). Even though vehicle dynamics can offer accurate hint on vehicle maneuvers, especially lane-change related maneuvers, it lacks robust and early detection. Thus, some researchers focused on using cues such as human behaviors in the inference. Doshi et al., used driver behavior cues (head movements and gaze position) to decide driver intention. Also, they analyzed the relative usefulness of different driver behaviors for determining driver intent (*12*).

Acknowledging that driver intention can be influenced by the external environment, many research studies have adopted an approach of combining multiple cues together to enhance the performance of intention inference models. Ding et al., proposed Comprehensive Decision Index (CDI) that reflects the influence of the surrounding traffic environment on drivers' lane-change decisions (*29*). Leonhardt et al., combined features describing the driving situation and the drivers' gaze behavior by means of data fusion, and the prediction of driving maneuvers is significantly improved (*30*). Xing et al., generated a vision-based bi-directional Recurrent Neural Network (RNN) with Long-Short Term Memory (LSTM) units' model that using both driver behavior and vehicle dynamics (*31*). Unlike previous research that predominantly focuses on front information, which may not adequately handle aborted maneuvers and changed intentions (such as aborted lane-change maneuvers), our paper considers both front and rear information. We believe that incorporating rear information can significantly aid in addressing false positive cases and improving the accuracy of intention recognition. Another noteworthy perspective is previous related research ignores the problem of user privacy issue when using the sensitive vehicle dynamic and driver behavior data. In this paper, drivers' privacy is considered through a FL-based framework.

## METHODOLOGY

This section provides an overview of the functioning of the PFL-LSTR framework in driver intention inference. It begins with a concise description of the research scope, outlining the specific areas of investigation. Subsequently, the workflow and algorithms of the proposed models are presented and elucidated in the following subsection.

### Driving maneuver analysis

According to Ou et al., the process of driving maneuvers can be conceptualized as a three-step model, which includes 1) observation action, 2) decision action, and 3) steering action (*14*). In particular, the observation action can be further categorized into two distinct types based on the specific driving maneuver:

1. Lane-keeping or lane-following related maneuvers involve the observation of the front of the vehicle. This observation period is referred to as the Frontal Observation Period (FOP).
2. Lane-change or turning related maneuvers require observing the surrounding environment. This observation period is known as the Surrounding Observation Period (SOP).

By distinguishing between these observation periods, one can better analyze different aspects of driving maneuvers. Drivers often observe surrounding traffic and assess safety factors before initiating steering actions, proceeding only when conditions are deemed favorable. Consequently, the focus of this paper lies in utilizing the information gathered during the observation action to infer drivers' intentions (human behavior data). It is important to note that during the Surrounding Observation Period (SOP) stage, drivers' intentions are significantly influenced by their surroundings. Therefore, the information pertaining to the surrounding environment becomes crucial within the inference framework. In light of this, the present study not only adopts the in-vehicle human driver behavior information but also adopts out-of-vehicle surrounding information (both vehicle front-view and rear-view information). This comprehensive approach allows us to incorporate the relevant details from both in-vehicle and out-vehicle (the front and rear of the vehicle), enabling a more comprehensive and accurate prediction of drivers' intentions.

Figure 2 illustrates the workflow of the proposed driver intention inference framework. The framework takes input information from three dash cams: front-view cam, rear-view cam, and in-cabin cam. These videos are processed to generate features using a pretrained ResNet50, which are then concatenated and fed as input to the PFL-LSTR framework. The LSTR model plays a vital role in predicting drivers' intentions, specifically lane-keep, left lane-change, and right lane-change. To achieve this, the LSTR encoder and decoder are trained using distinct memory mechanisms, enabling the model to capture different temporal dependencies and patterns in the driving behavior. LSTR encoder is trained through long memory, but LSTR decoder is trained through work memory (short memory). In the proposed framework, during each communication round, the LSTR decoder is trained individually by each local user, resulting in a unique LSTR decoder tailored to their specific local dataset. On the other hand, the LSTR encoder is trained using a weighted average approach, where selected local users contribute to the training process. This collective training enables the LSTR encoder to learn a common representation from the input of the chosen local users. By combining the personalized LSTR decoder and the shared knowledge learned by the LSTR encoder, the framework achieves a powerful combination of personalized and collective prediction capabilities. The technical details are introduced in the following two subsections.

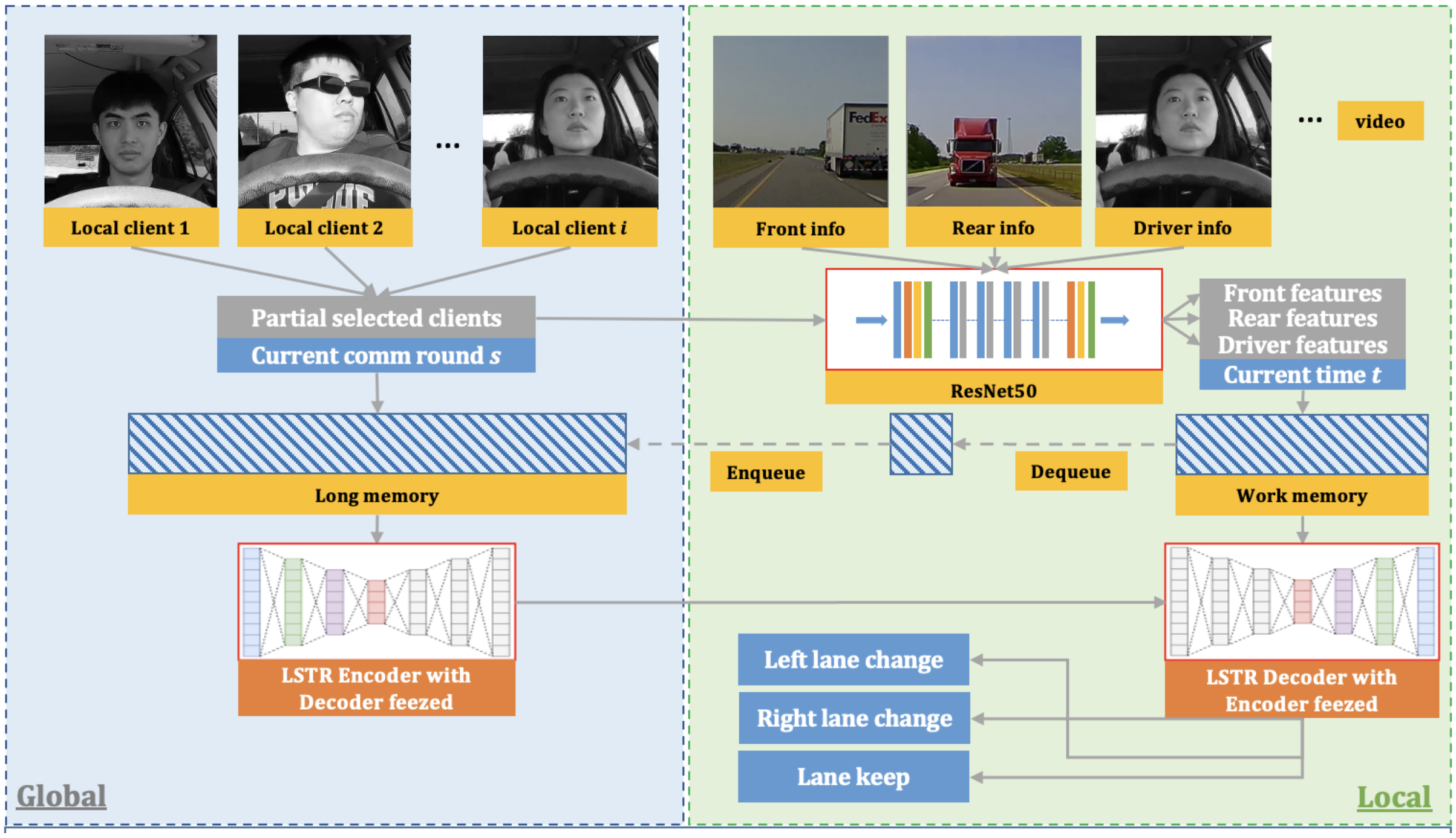

**Figure 2. Driver intention inference framework (gray represents data, blue represents variable, orange and yellow represents algorithms and notations)**

**Local fine-tuning based PFL**

Personalized Federated Learning (PFL) was used in this study as the main architecture for intention prediction. PFL offers the dual benefits of preserving user privacy while also incorporating personalized modeling techniques. By leveraging PFL, sensitive user data can be protected while tailoring the prediction models to account for the inherent heterogeneity in driving styles among different individuals. This personalized approach enhances the accuracy and relevance of our intention predictions, ensuring both privacy preservation and personalized modeling are effectively addressed within the framework.

Moreover, local fine-tuning serves as a classic method to introduce personalization to federated learning. Through this approach, the PFL framework efficiently learns both a common representation, shared among all local user data, and unique representations specific to each local user. In contrast to the traditional FedAvg model, where multiple local updates in a heterogeneous setting can lead each client away from the best averaged representation, our approach adopts a different strategy. In our proposed method, each client or local user conducts several local updates during each communication round. This approach offers several advantages, including the ability to learn an individualized model for each client, ultimately enhancing the overall performance of the model. This approach proves advantageous as it enables the learning of an individual model, enhancing the overall performance and mitigating the negative effects caused by heterogeneity. By allowing for multiple local updates, our method facilitates personalized learning and contributes to improved model accuracy within the federated learning framework. Ensuring generalization to new clients or local users is a pivotal task in federated learning. Similar to adaptability, the ability to quickly generalize to a new local user is essential for the framework to be applicable in real-world scenarios. By using local fine-tuned PFL, a representation already available, thus new local user only need to have a few epochs to train the unique representation.

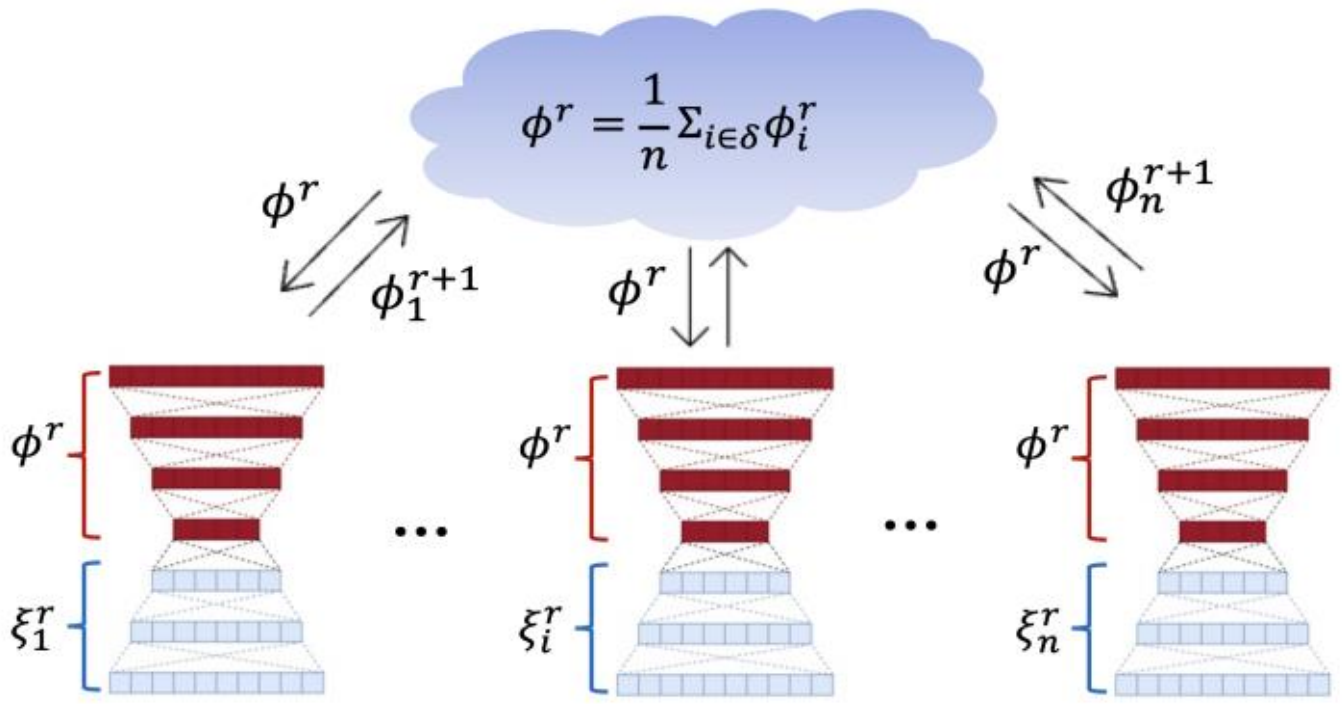

**Figure 3. PFL with global representation $\phi$ updated by selected server and unique representation $\xi_i$ learnt by each client**

As shown in Figure 3, the local clients and global server try to learn a common representation together as $\phi$, while each local client learn their unique representations as $\xi$. During the client update stage in federated learning, which occurs during local training, each client $i$ undergoes $\tau$ local gradient-based updates using the current global representation $\phi^r$. In other words, each client performs a series of updates based on its local dataset to refine its own local representation. Once the local representations are updated, a subset of clients selected from the set $\delta$ proceed to update the global representation. These selected clients perform additional steps of updating each global representation $\phi_i$ and then transmit these updates to the server. After receiving the updates from the selected clients, the server generates the new global representation by computing a weighted average of all the updated global representations, which can be represented as:

$$\min_{\phi\in\Phi}\frac{1}{n}\sum_{i=1}^{n}\min_{\xi\in\Xi} f_i(\xi_i, \phi)$$

This global representation then serves as the basis for the subsequent rounds of training, allowing the federated learning system to continue improving and refining its model based on the contributions from the distributed clients.

**LSTR embedded PFL for driving intention detection**

In our proposed approach, we utilize drivers' observation action and the information of surrounding traffic to accurately deduce their intentions. The LSTR model is employed for this task, offering efficient handling of long videos. Additionally, we address concerns regarding user privacy, handle diverse data, and enhance the framework's adaptability by incorporating local-fine tuning PFL. This combination ensures precise intention prediction and boosts the overall framework performance and adaptability. By integrating these elements, our method provides a comprehensive solution to address the intricacies of driver intention inference and fosters a safer and more streamlined transportation environment.

*LSTR for driver intention inference*

LSTR (Long-Short Term Transformer) stands out as a highly suitable option for driver intention inference, primarily because of its exceptional capabilities in online temporal modeling of extended video sequences. In the context of inferring driver intentions, the importance of causal data processing cannot be overstated because it enables predictions based solely on the current and past information without any knowledge of future frames. LSTR exhibits excellent potential precisely in this area, as it is tailored specifically for online action detection and this makes it a perfect fit for real-time applications like driver intention prediction. Its ability to process data causally and handle extended video sequences efficiently makes it a promising and powerful choice for accurately deducing driver intentions in various driving scenarios.

Figure 4 showcases the division of history videos into long-term memory and short-term memory within the LSTR model's encoder-decoder architecture. The LSTR encoder efficiently compresses and abstracts the long-term memory into a fixed-length latent representation. On the other hand, the LSTR decoder operates on a short window of transient frames, leveraging self-attention and cross-attention operations on the token embeddings extracted from the LSTR encoder. The data flow of long- and short-term memories is depicted with dashed arrows, following the first-in-first-out (FIFO) logic. To elaborate, the short memory, also known as work memory, stores a limited number of currently observed frames in a FIFO queue with $m_s$, slots. As a frame surpasses the $m_s$, threshold, it moves into the long memory queue, which also operates as a FIFO queue but with $m_l$ slots. The long-term memory serves as input to the LSTR encoder, while the short-term memory acts as input to the LSTR decoder.

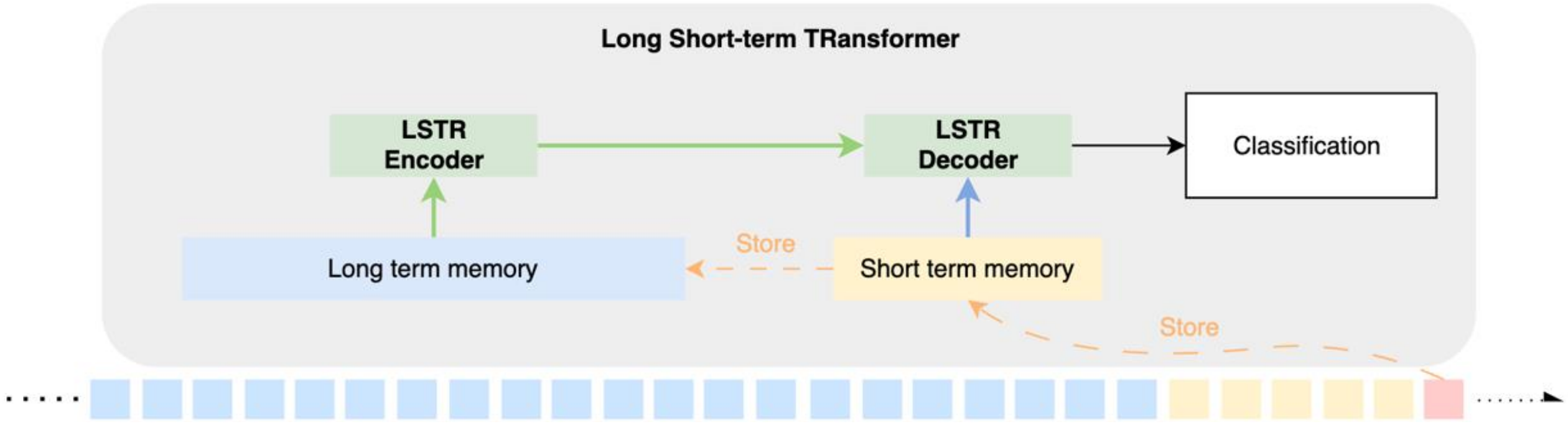

**Figure 4. Workflow of Long Short-term TRansformer**

*Personalized Federated Learning based LSTR*

The distinctive two-memory structure of the LSTR model facilitates its seamless integration with personalized federated learning, particularly in the context of fine-tuning. Leveraging this approach, the long-memory section can be effectively trained and updated on the global server, capturing common representations and characteristics from the aggregated driving data. This global training process enables the model to identify patterns applicable across multiple drivers, promoting a generalized understanding of driving behavior.

Conversely, the short-memory section undergoes local training on each client's device, allowing the model to learn and adapt to the unique driving experiences and behaviors of individual drivers. Fine-tuning the short-memory section on the client's device ensures that the model accommodates the idiosyncrasies of each driver, providing personalized and context-specific predictions for their driving intentions. This combination of global and local training enhances the overall performance of the model and addresses the challenges of data heterogeneity, ensuring that the framework is well-suited for real-world deployment in diverse driving scenarios. During the training of the PFL-LSTR model, a classic FedAvg is used in the first round to initialize the global model. Then for each communication round, selected clients update the LSTR encoder using long-memory by weighted average, then each client updates the LSTR decoder using short-memory. Algorithm 1 presents the PFL-LSTR training process:

**Algorithm 1: Personalized Federated Learning - Long Short-term Transformer model**

Initialize global model weights: $\phi_0$
**for** communication round $t = 0$, **do**
    global model $\phi_0$distributed to the local client with total data sample $N_k$
    for local model with weight $\phi_0^i$ $i = 1, 2, \ldots, k$ **do**
    get local data samples: $N_i$
    $\phi_1^i$ = local server train ($\phi_0$, epoch: $e$)
global model weights $\phi_1 = \sum_{i=1}^{k} \phi_1^i \frac{N_i}{N_k}$
Using the initialized FedAvg weights: $\phi_1$
**for** communication round $t$ = 1, 2, …, $T$ **do**
    global model $\phi_t$ is distributed to the local client
    $\xi_{t+1}^i$ = local server train ($\xi_t$, local epoch:$e_l$)
    **for** selected client in $S_k$ **do**
        get local data samples: $N_i$
        $\phi_{t+1}^i$ = selected local server train ($\phi_t$, epoch: $e$)
    $N_{s_k}$= total selected data samples
    global model weights $\phi_{t+1} = \sum_{i=1}^{k} \phi_{t+1}^i \frac{N_i}{N_{S_k}}$
each personalized model weight $\omega_{t+1} = f(\xi_t, \phi_t)$

# EXPERIMENTS

## Data collection scenarios and data preprocessing

This study utilizes real-world human drivers' driving data for analysis and investigation. Three different dash cameras are mounted on a vehicle with front-view, rear-view and driver-view captured (as shown in Figure 5(a) to Figure 5(c)). The driving data is exclusively collected on Interstate Highway (e.g., I-65, the driving route is shown in Figure 5(d)), which is renowned for its high traffic volume and relatively high traveling speeds. As shown in Figure 5 (e), most of the time the vehicle maintains traveling speed around 80 mph. For this study, we have gathered driving data from three different drivers, resulting in a total driving length of approximately 4 hours. The collected driving videos can be divided based on the maneuvers' type into three distinct driving scenarios, each characterized by different observation actions:

1. Lane-change with surrounding observation actions: In this scenario, drivers exhibit evident body movement, head movement, and changes in gaze position while preparing for a lane-change maneuver.
2. Lane-keep with front observation actions: This scenario involves drivers maintaining their lane with stable body movement or head movement, while keeping their gaze fixed on the front.
3. Lane-keep but with surrounding observation actions: Here, drivers show observation actions similar to lane-change maneuvers, even though they do not intend to change lanes. These cases can lead to false positive predictions, where drivers display observation actions without engaging in actual lane-changes or turns. This behavior is often associated with checking dashboard instruments, using navigation tools, or communicating with passengers, and it can result in false alarms and negative effects on the driver.

Figure 5 (f) illustrates that drivers with diverse driving habits exhibit varying false positive rates, indicating significant individual variability in this aspect. Addressing false positive cases is a key objective of this study, as it presents a critical challenge in driver intention inference. Remarkably, prior research has not specifically tackled this issue, making our investigation all the more significant. By focusing on mitigating false positive predictions, we aim to enhance the accuracy and reliability of driver intention inference algorithms, thereby improving overall transportation safety and driver assistance systems.

**Training and testing settings**

In the model training section, we employed our proposed model and conducted training over 100 communication rounds. During each communication round, the LSTR decoder is updated with 5 local epochs to learn from each client's specific driving behavior. The LSTR encoder is updated through a weighted average approach with a learning rate of $10^{-6}$ by selected clients using 1 epoch each to capture common representations and characteristics from their driving data. This ensures an efficient and coherent update of the global model while incorporating valuable insights from multiple clients. To optimize the model's performance, we carefully selected the work memory length to be 3 seconds, as this time range effectively captures most of the observation actions. The long memory length is set to be 12 seconds, because using longer time intervals may introduce additional noise and disturbances that could potentially impact the model performance. Striking the right balance between the length of long memory and its impact on performance is crucial in achieving accurate and reliable driver intention inference.

Furthermore, the proposed model is benchmarked against other baseline models to evaluate its performance: the classic FedAvg model (without personalization) and the local LSTR model. Additionally, we conduct experiments with variations of the proposed model, one with the inclusion of a rear-view camera and one without. This comparison allows us to assess the impact of using rear-view camera data in reducing false positive predictions, thereby improving the overall accuracy of driver intention inference.

- FedAvg is a classic Federated Learning (FL) model, incorporating LSTR for driver intention prediction. It updates the global model based on data from all clients in 100

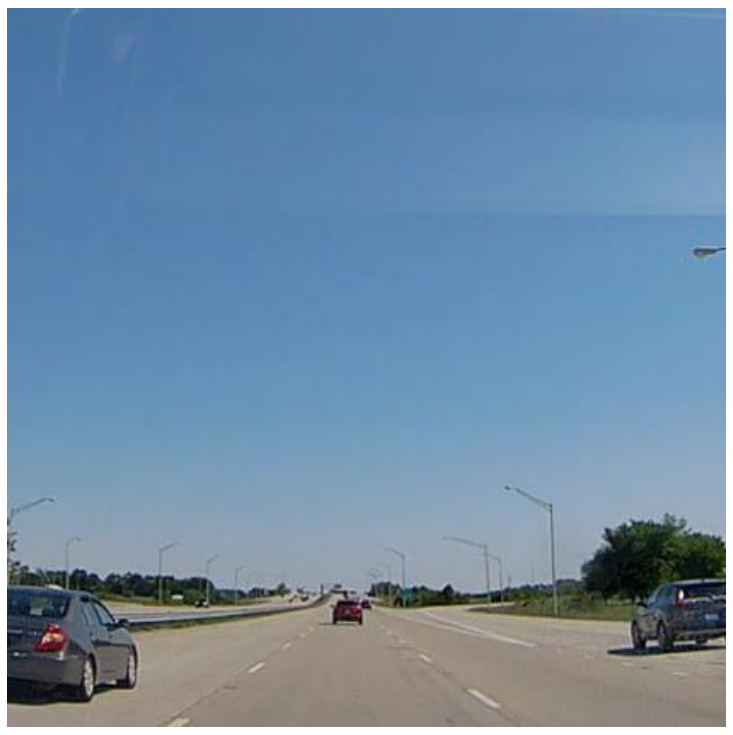

(a) Front-view

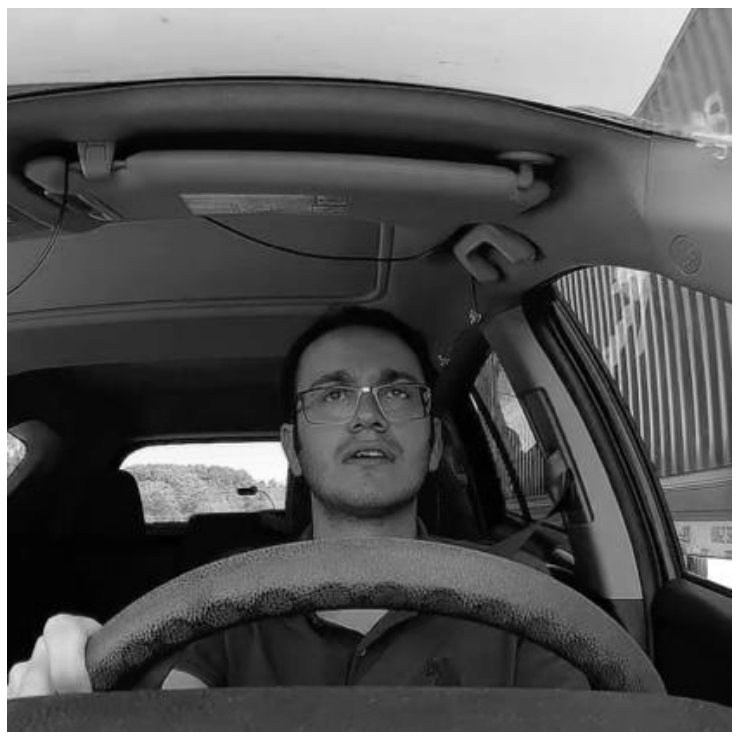

(b) Driver-view

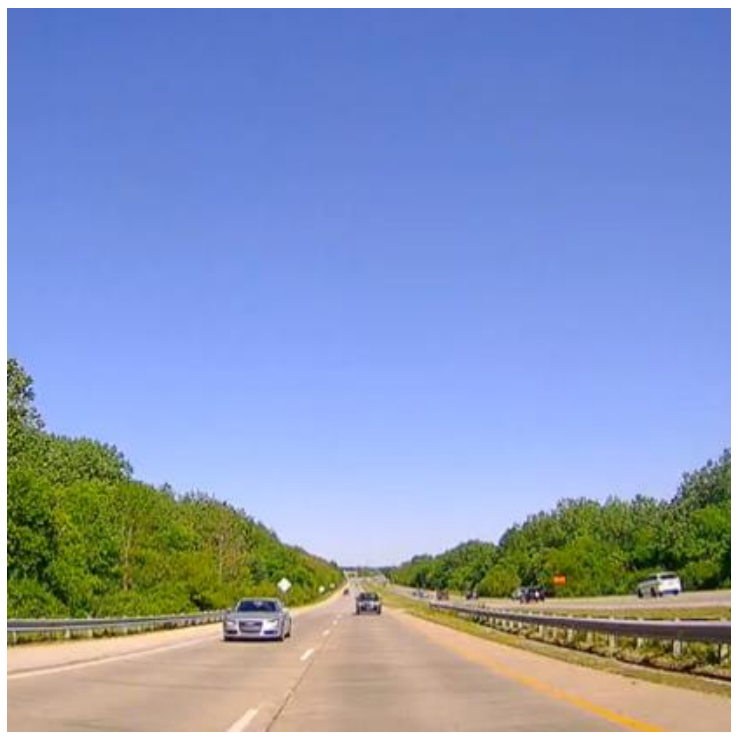

(c) Rear-vew

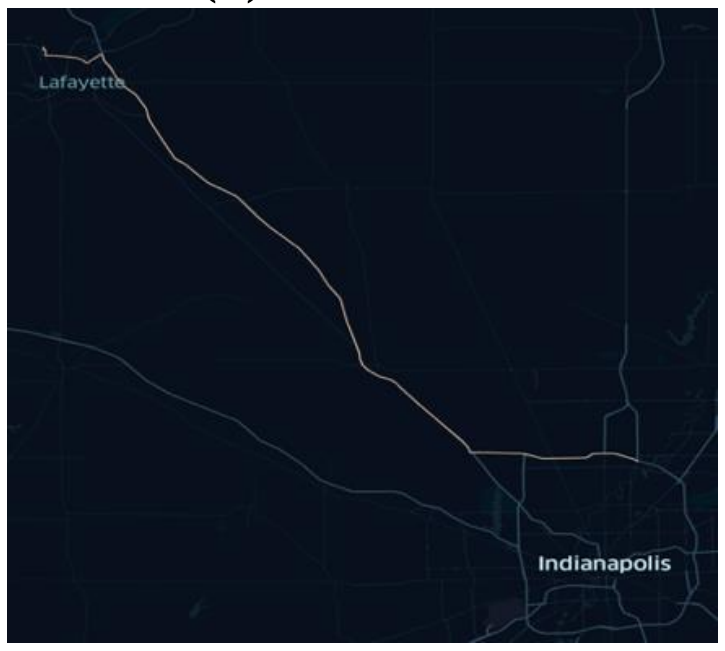

(d) Data collect route

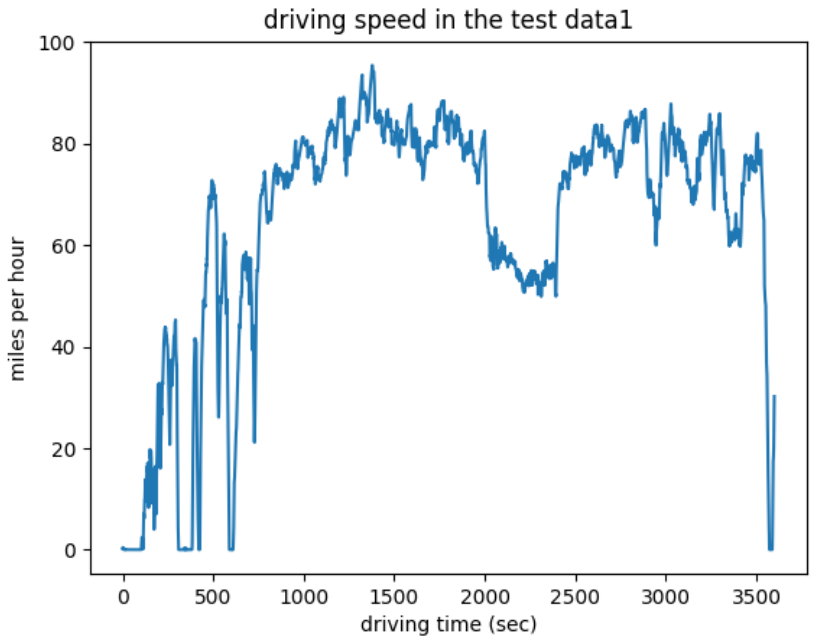

(e) Traveling speed during data collection

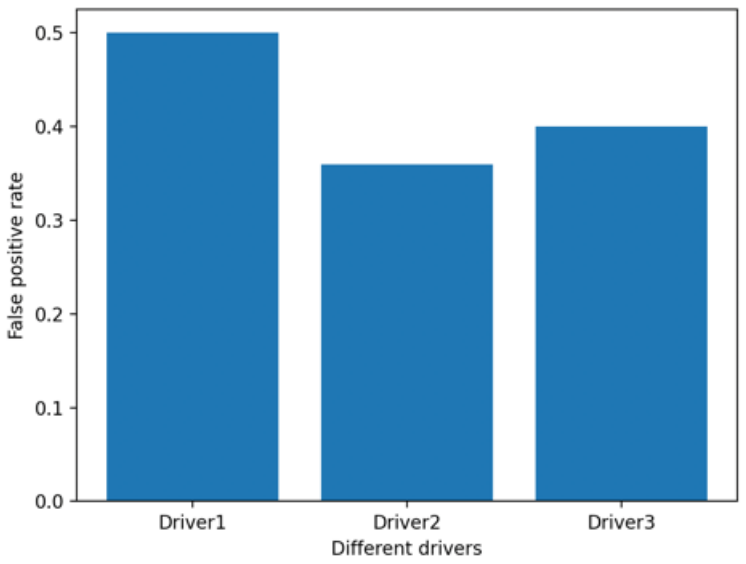

(f) False positive rates of different drivers

**Figure 5. Information captured by three-view dash cams**

communication rounds. The learning rate is set to $10^{-7}$, and batch size and memory length match the proposed model.

- The local LSTR models are trained on local servers without sharing and aggregating parameters to the cloud. These models are highly personalized and adapt to specific local data. The settings are the same as the proposed model and FedAvg, but the total epochs are set to 1000.
- PFL-LSTR without rear-view information shares the same model structure and settings as the proposed model. However, the input information includes only front-view and in-cabin view data, excluding rear-view information.

## RESULTS & DISCUSSION

### Personalization: the adaptability of different models

Personalization is indeed a critical factor in driving intention inference, as each driver's behavior can vary significantly based on their driving style, experience, and preferences. By customizing the intention inference model to each individual driver, we can achieve a more precise and context-specific prediction of their driving intentions. In this study, we compare the performance of our proposed PFL-LSTR model and the classic FedAvg model in terms of model accuracy during training and testing on diverse local datasets. The accuracy metric serves as an indicator of the model's ability to adapt effectively to individual drivers and their unique driving scenarios. Higher accuracy indicates a better level of adaptability, making personalized inference more reliable and valuable for real-world applications.

Figure 6 (a) illustrates the training results of three different models on the testing set, showcasing their varying inference precision. For driver 1's dataset, which contains a larger number of data points, both FedAvg and our proposed PFL-LSTR model demonstrate good performance. However, the proposed PFL-LSTR model achieves a faster convergence to a relatively high precision level. In contrast, for driver 2's dataset with fewer data points, FedAvg's precision is negatively impacted, while PFL-LSTR maintains stable performance. The most intriguing observation emerges from driver 3's dataset, where FedAvg struggles to learn from the driving behavior, resulting in precision below 20%. Conversely, PFL-LSTR attains an impressive precision of 70% on the same dataset.

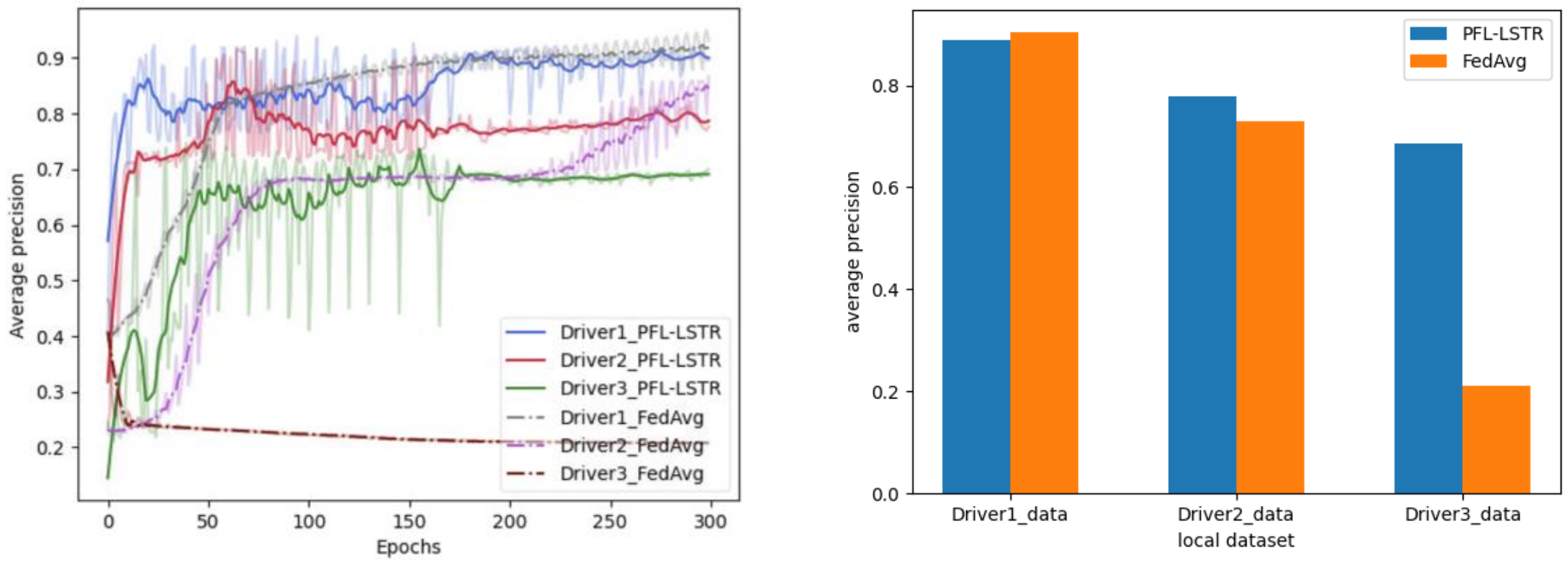

(a) Precision curve of the two models on different lcoal dataset during training

(b) Average precision of the two models after converge on different local dataset

**Figure 6. PFL-LSTR vs FedAvg precision on different local users' data**

Figure 6 (b) further confirms the promising results on the testing dataset, showcasing that PFL-LSTR consistently maintains higher average precision even after model convergence. These findings highlight the superior adaptability of the proposed PFL-LSTR model, particularly for local clients with heterogeneous data distributions. The significant benefits of personalization become evident, enabling the model to effectively handle diverse driving behaviors and data patterns among individual drivers.

**Training efficiency: loss decreasing rate for different models**

Training efficiency is crucial, particularly in a network that handles multiple local clients, as it directly impacts the safety and effectiveness of the traffic system. By ensuring high efficiency learning and service, we can enhance the real-time capability of the model and provide timely and accurate predictions for driver intentions. In the training stage, PFL-LSTR employs personalized model training for each local client through a combination of local fine-tuning PFL and LSTR sections. This approach enables the model to adapt to different drivers with diverse driving habits, resulting in more context-specific predictions. Additionally, the global model is trained to learn common representations from all clients, minimizing the distances between these parts and facilitating faster convergence during the training process.

Indeed, training efficiency can be significantly enhanced through the utilization of federated learning (FL). FL offers the advantage of accessing a large volume of data from diverse clients without centralizing it in a single location. This decentralized data access enables the model to benefit from a more comprehensive and representative training dataset, resulting in improved generalization and performance. Additionally, during the FL process, models from different clients collaborate by exchanging information through model updates. This knowledge transfer fosters the sharing of valuable insights and diverse experiences, allowing the model to learn from different perspectives. As a result, the model's learning process is accelerated, and convergence is achieved more rapidly. By leveraging the collaborative nature of federated learning, the proposed PFL-LSTR framework efficiently harnesses the collective intelligence of multiple clients, leading to a powerful and adaptive driver intention inference system with reduced computational costs and enhanced training speed.

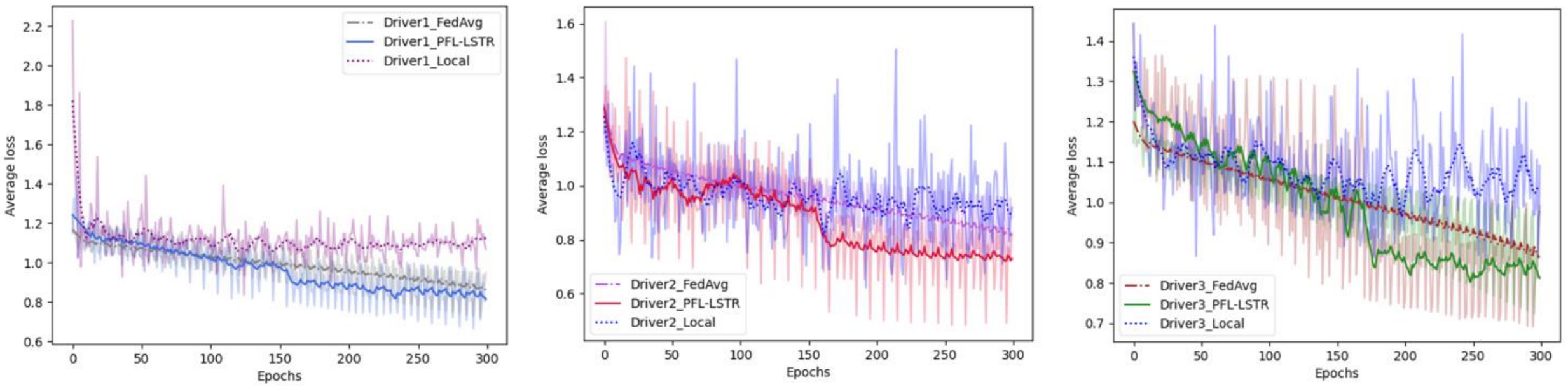

(a) PFL-LSTR, FedAvg, and Local training efficiency on Driver 1 dataset

(b) PFL-LSTR, FedAvg, and Local training efficiency on Driver 2 dataset

(c) PFL-LSTR, FedAvg, and Local training efficiency on Driver 3 dataset

**Figure 7. PFL-LSTR, FedAvg and local model training efficiency on different local users' data**

Figure 7(a) to Figure 7(c) clearly demonstrate the superior training efficiency of both FedAvg and PFL-LSTR over the local model. Notably, the advantage becomes particularly evident at the initial stages of the training process. In the case of specific datasets, such as the datasets for driver 2 and driver 3, PFL-LSTR exhibits superior performance compared to FedAvg. The reason behind this lies in FedAvg, which trains a single global model encompassing data from all drivers. Consequently, the substantial diversity among the driving behaviors of different drivers leads to varying convergence times for certain local clients' data.

In contrast, PFL-LSTR's personalized approach allows it to effectively adapt to individual drivers with diverse driving habits and behaviors. By leveraging local-fine tuning and long short-term memory sections, the PFL-LSTR model tailors itself to the unique characteristics of each driver's data. As a result, PFL-LSTR achieves quicker convergence, even when faced with heterogeneous data distributions among the local clients. This personalized adaptation is a crucial factor contributing to the superior training efficiency of PFL-LSTR compared to FedAvg and the local model.

**Precision for different maneuvers and false positive cases**

During the testing stage, we evaluate our model's performance across various driving maneuvers, including lane-change maneuvers (left lane-change and right lane-change) and lane-keep maneuvers. In these scenarios, false positive cases are inevitable, such as lane-keep maneuvers with evident head movement, which may be incorrectly identified as lane-change maneuvers. Our analysis highlights the significance of rear-view information in the context of vehicle surroundings, an aspect that has not been adequately emphasized in previous research. In our investigation of the impact of the rear-view camera on intention inference performance, we conducted tests on models with and without rear-view information using the testing dataset.

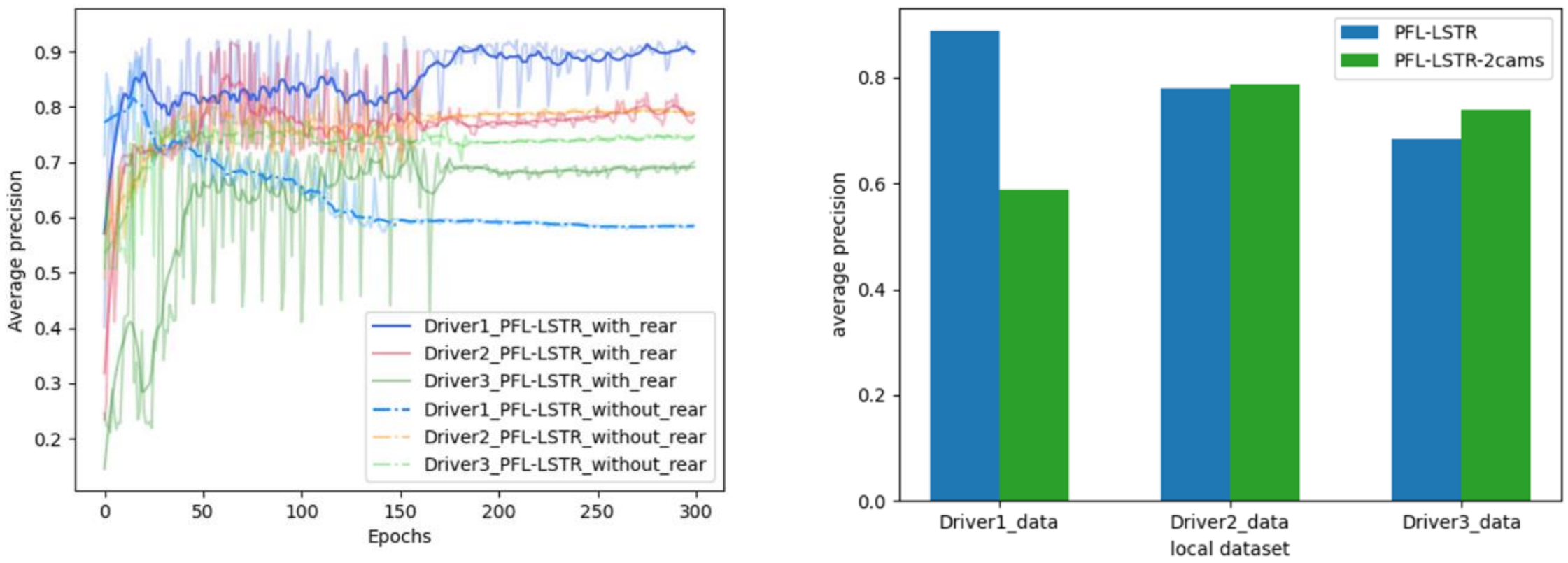

(a) Precision curve of the two models on different lcoal dataset during training

(b) Average precision of the two models after converge on different local dataset

**Figure 8. PFL-LSTRvs PFL-LSTR-2cams precision on different local users' data**

In Figure 8(a), we observe that the absence of rear-view information has a considerable impact on the precision of the model, particularly for driver 1's dataset during the training stage. The precision decreases by approximately 30% due to a significant number of false positive cases, where lane-keep scenarios are misidentified as lane-change maneuvers, likely because of evident head movement and other gestures. The lack of rear-view information hampers the model's ability to accurately infer the driver's intention, resulting in a substantial reduction in precision. However, for driver 2 and driver 3, who have a lower occurrence of false positive cases (approximately 30%), the effect of the rear-view camera on intention inference appears to be less pronounced. In these instances, the absence of rear-view information does not significantly impact the model's performance, suggesting that the model can rely on other observable cues to make accurate predictions. This underscores the importance of considering individual driving behavior and data distribution when evaluating the relevance of specific information sources in the intention inference process.

On the testing dataset, the precision of inference three different maneuvers intention are analyzed. FedAvg and PFL-LSTR-2cams are compare with the proposed model. As shown in Figure 9(a) and Table 1, the comparison between PFL-LSTR and FedAvg on the testing set with diverse scenarios clearly highlights the superior performance of PFL-LSTR. On average, PFL-LSTR consistently outperforms FedAvg across all different local datasets. Particularly noteworthy is the stable and accurate performance of PFL-LSTR in inferring lane-keeping maneuvers, where includes a lot of false positive cases. On the contrary, FedAvg struggles to make accurate inferences for lane-keeping maneuvers, often resulting in incorrect predictions. This discrepancy arises from the presence of false positive cases in each individual client's dataset. FedAvg, lacking personalization capabilities, tends to make erroneous inferences in the face of such diverse and context-specific driving behaviors. In contrast, the personalized approach of PFL-LSTR, with its fine-tuning mechanism and long short-term memory sections, enables it to adapt to the unique characteristics of each driver's data. As a result, PFL-LSTR delivers more accurate and context-specific predictions, making it a superior choice for driver intention inference in various driving scenarios.

**Table 1. PFL-LSTR, FedAvg, and PFL-LSTR-2cams performance on various dataset**

| | Driver1 | | | Driver2 | | | Driver3 | | |
|---|---|---|---|---|---|---|---|---|---|
| Models | PFL-LSTR | FedAvg | PFL-LSTR-2cams | PFL-LSTR | FedAvg | PFL-LSTR-2cams | PFL-LSTR | FedAvg | PFL-LSTR-2cams |
| Lane-keep | 0.78 | 0.32 | 0.24 | 0.95 | 0.95 | 0.95 | 0.98 | 0.96 | 0.52 |
| Left lane-change | 0.94 | 0.63 | 0.90 | 0.79 | 0.66 | 0.34 | 0.98 | 0.95 | 0.95 |
| Right lane-change | 0.98 | 0.21 | 0.95 | 0.90 | 0.99 | 0.99 | 0.25 | 0.27 | 0.29 |

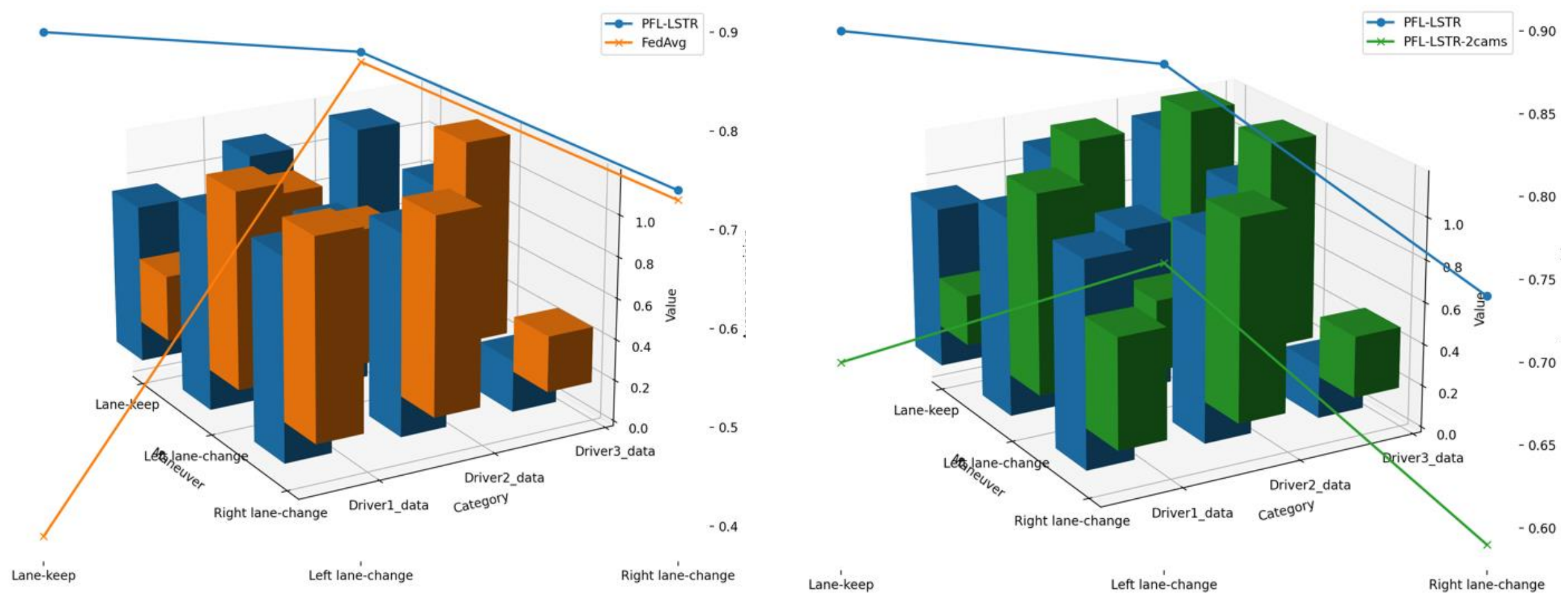

(a) PFL-LSTR vs FedAvg (b) PFL-LSTR vs PFL-LSTR-2cams

**Figure 9. Precision of various maneuvers on testing dataset**

Moreover, as shown in Figure 9(b), on the testing dataset, the proposed PFL-LSTR significantly outperforms the PFL-LSTR-2cams. The performance of the FedAvg is superior to that of PFL-LSTR without the rear-view information. In other words, the rear-view information can be really important in enhancing the intention inference accuracy and avoid false positive cases. From Table 1, it can be observed that the average precision of driver1's lane-keep maneuver is only 24% through using the two cameras' data, while the same maneuver precision can be enlarged to 78% when there is both personalization and rear-view information provided. More importantly, as shown in Table 1, lack of rear-view information could cause false negatives. For example, during a lane-change maneuver, the driver may check the rear-view mirror to assess the position and speed of vehicles behind them before executing the lane change. If the model does not have access to this rear-view information, it might not be able to accurately predict the driver's intention to change lanes, leading to a false negative case where the model fails to identify the lane change.

## CONCLUSION

This study demonstrates the significance of driver intention inference, and particularly considers fusion various information (including both in-vehicle and out-vehicle), personalization and user privacy. The proposed PFL-LSTR model not only considers user privacy but also showcases superior adaptability and inference precision, particularly when faced with heterogeneous data distributions among local clients. By employing local-fine tuning based PFL and LSTR with long short-term memory sections, the PFL-LSTR model tailors itself by combining both the common representation and unique characteristics of each drivers' data, resulting more accurate and context-specific intention inference.

At both the training and testing stages, the proposed PFL-LSTR was compared with other baseline models: FedAvg, local model, and PFL-LSTR without rear information. Compared with classic FedAvg, PFL-LSTR exhibits superior adaptability to various local dataset with higher inference precision during training. In the local dataset with heterogeneous data distribution, PFL-LSTR outperforms the FedAvg by more than 50%. Training efficiency is also a critical factor, and the federated learning approach contributes to enhanced efficiency by leveraging a large volume of data from multiple clients without centralizing it. The collaborative nature of federated learning allows models from different clients to exchange information, leading to quicker convergence and improved performance. The importance of rear-view information in driving intention inference is highlighted, as the study suggests that its absence could lead to decreased precision and even false negative cases. Rear-view information proves particularly crucial in scenarios with a higher occurrence of false positive cases, where it helps generate more accurate predictions and avoids misclassifications. In both training and testing stage, PFL-LSTR showcases stable performance on the local dataset with high percentage of false positive cases, and the inference precision is 46% and 54% higher than the classic FedAvg and PFL-LSTR without rear-view information. Thus, we show that rear-view information can be really important in driver intention inference to decrease the false positive rate.

Overall, the proposed PFL-LSTR model offers a powerful and adaptive solution for driver intention inference, ensuring accurate and context-specific predictions while considering the unique characteristics of each driver's behavior and driving scenario. Moreover, by considering comprehensive in-vehicle and out-vehicle information, we proved that false positive cases can be decreased by combining personalization and comprehensive information together.

## ACKNOWLEDGMENTS

This work was supported by Purdue University's Center for Connected and Automated Transportation (CCAT), a part of the larger CCAT consortium, a USDOT Region 5 University Transportation Center funded by the U.S. Department of Transportation, Award #69A3551747105. The contents of this paper reflect the views of the authors, who are responsible for the facts and the accuracy of the data presented herein, and do not necessarily reflect the official views or policies of the sponsoring organization.

## AUTHOR CONTRIBUTIONS

The authors confirm contribution to the paper as follows: all authors contributed to all sections. All authors reviewed the results and approved the final version of the manuscript.